%% file: root.tex
\documentclass[letterpaper, 10 pt, conference]{ieeeconf}  % Comment this line out if you need a4paper

\IEEEoverridecommandlockouts                              % This command is only needed if 
\input{include/packages}

\title{\LARGE \bf
RTAW: An Attention Inspired Reinforcement Learning Method for Multi-Robot Task Allocation in Warehouse Environments
}

\author{Aakriti Agrawal$^{1}$, Amrit Singh Bedi$^{1}$, and  Dinesh Manocha$^{1}$ % <-this % stops a space
\thanks{$^{1}$The Department of Computer Science, University of Maryland, College Park, MD, USA {\tt\small \{agrawal5,amritbd,dmanocha\}@umd.edu}}}

\begin{document}
\maketitle
\thispagestyle{empty}
\pagestyle{empty}

%%%%%%%%%%%%%%%%%%%%%%%%%%%%%%%%%%%%%%%%%%%%%%%%%%%%%%%%%%%%%%%%%%%%%%%%%%%%%%%%
\begin{abstract}

We present a novel reinforcement learning based algorithm for multi-robot task allocation problem in warehouse environments. We formulate it as a Markov Decision Process and solve via a novel deep multi-agent reinforcement learning method (called RTAW) with attention inspired policy architecture. Hence, our proposed policy network uses global embeddings that are independent of the number of robots/tasks. We utilize proximal policy optimization algorithm for training and use a carefully designed reward to obtain a converged policy. The converged policy ensures cooperation among different robots to minimize total travel delay (TTD) which ultimately improves the makespan for a sufficiently large task-list. In our extensive experiments, we compare the performance of our RTAW algorithm to state of the art methods such as myopic pickup distance minimization (greedy) and regret based baselines on different navigation schemes. We show an improvement of upto 14\% (25-1000 seconds) in TTD on scenarios with hundreds or thousands of tasks for different challenging warehouse layouts and task generation schemes. We also demonstrate the scalability of our approach by showing performance with up to $1000$ robots in simulations.

\end{abstract}
% \keywords{Mult-robot task allocation, Reinforcement Learning, Large Warehouses, Optimisation.} 

%%%%%%%%%%%%%%%%%%%%%%%%%%%%%%%%%%%%%%%%%%%%%%%%%%%%%%%%%%%%%%%%%%%%%%%%%%%%%%%%
\input{Introduction.tex}

\input{algo.tex}
\input{Simulations.tex}

\input{conclusions.tex}

%%%%%%%%%%%%%%%%%%%%%%%%%%%%%%%%%%%%%%%%%%%%%%%%%%%%%%%%%%%%%%%%%%%%%%%%%%%%%%%%

%References are important to the reader; therefore, each citation must be complete and correct. If at all possible, references should be commonly available publications.

\bibliographystyle{IEEEtran}
\bibliography{example}

\input{appendix.tex}

\end{document}

%% file: include/packages.tex
\usepackage{amsmath}
\usepackage{amssymb}
\usepackage{graphicx}
\usepackage{cite}
\usepackage{booktabs}

\usepackage{amsthm}

\usepackage[dvipsnames]{xcolor}
\usepackage{subcaption}
\usepackage{wrapfig}
\usepackage{multirow}
\usepackage{mathtools}
\usepackage{empheq}
\usepackage{color}
\usepackage{bbm}
\usepackage{xspace}
\let\labelindent\relax
\usepackage[shortlabels]{enumitem}

\usepackage{algorithmicx}
\usepackage{algorithm}
\usepackage[]{algpseudocode}

\usepackage{fontawesome}

\usepackage{soul}
\usepackage{comment}

\usepackage[hyphens]{url}
\usepackage[pdftex, colorlinks=true, linkcolor=blue, citecolor = black, urlcolor = black, filecolor=black , pagebackref=true, hypertexnames=false]{hyperref}
\usepackage{cleveref}
\crefname{section}{Sec.}{Secs.}
\Crefname{section}{Section}{Sections}
\Crefname{table}{Table}{Tables}
\crefname{table}{Tab.}{Tabs.}
\crefname{figure}{Fig.}{Figs.}

%% file: Introduction.tex
\section{Introduction}
Recent advancements in artificial intelligence (AI) and robotics have contributed to the widespread use of automation in warehouse environments. A recent case study by Honeywell\footnote{\url{https://tinyurl.com/344p5e4y}} indicated more than a 50\% rise in e-commerce sales after COVID-19 amidst existence of human labor shortage, thus proving the need of automating fulfillment and distribution centers/warehouses. Warehouse automation, widely being used by online retailers like Amazon, Alibaba, etc, can be performed by using multiple robots and a decision-making system. We are considering the task of picking and delivering a parcel from point A to point B \cite{warehouse_comb}. By designing an appropriate multi-robot decision-making system, we can improve the task efficiency and throughput. The problem of designing such a system can be further divided into two parts: \emph{task allocation}~\cite{taxonomy} and \emph{robot navigation}~\cite{mapd_review}. Generally, greedy approaches are used to solve task allocation and multi-agent planning for navigation \cite{cbs_ta,usc}. 

We mainly deal with the problem of task allocation in multi-robot settings, which is popularly known as multi-robot task allocation (MRTA). This problem is widely studied in the literature using techniques from combinatorial optimization~\cite{taxonomy}. The problem of computing the optimal solution is NP-hard in general, and most practical solutions are based on greedy techniques~\cite{murdochs,usc}. A comprehensive survey of MRTA problems is given in~\cite{task_survey}, which divides the problem into multiple categories depending on the robot types, task types, and task assignment to robots. 

In this paper, we address the problem corresponding to the ST-SR-IA category, where we consider single-task robots (ST) which execute only a single task at any given time. All tasks are single robot tasks (SR) and task assignment is instantaneous (IA) where the task is instantaneously allocated at each decision timestep. This class of problems can be solved in polynomial time at each decision timestep~\cite{taxonomy}. Prior methods focus on developing a deterministic solution for task allocation for a given set of tasks and robots~\cite{comb_opti}. We are interested in a sequential-lifelong version of the problem where new tasks are continuously generated and task is allocated to a robot as soon as it becomes available. Our goal is to maximize the sum of utilities over time, that is, assign tasks at each instant in such a way that the total time taken to execute a specific number of tasks \em{makespan} as well as extra travel time \em{TTD} is minimized. 

The existing state-of-the-art task allocation methods are mainly designed for multi-agent pickup and delivery (MAPD) and are primarily coupled with grid based navigation schemes, which are mostly designed for centralized systems~\cite{usc}. Our method when combined with decentralised navigation algorithms can give rise to decentralised task-allocation and navigation with obstacle avoidance \cite{aakriti_iros}. But the current practical warehouses uses only grid based navigation. Thus, this work focuses on the commonly occurring system in warehouses which is based on centralized task allocation and navigation.

\textbf{Main Results:} The novel components of our work include:

\begin{enumerate}
\item We formulate (first time in literature) the MRTA problem in warehouse settings as a reinforcement learning problem. This require careful design of state, rewards, and actions for the problem to utilize any existing method to solve the problem. 
\item  The selection of reward for MRTA problem is quite challenging and we propose to use total travel delay (TTD) as the reward after evaluating different rewards like task length, total time of robot travel, etc.  Interestingly, only the TTD based rewards exhibit converging behavior. 
\item We propose a novel policy architecture that can handle a variable number of robots and tasks easily by forming useful global encoded representation using a mechanism inspired from attention. This is important because we can utilize the similar model to accommodate changes in the robot number during the execution process.  with different number of robots. This also makes RTAW scalable to thousands of robots (Sec.~\ref{scalability}) in our simulations with constant space complexity and linear time complexity with respect to number of agents and tasks (cf. Appendix \ref{time_space}).
% (cf. \cite{icra_supp} Appendix VII).
\item We show extensive evaluations and compare against the state-of-the-art greedy baseline called myopic pickup distance minimization (MPDM) and regret-based task-selection (RBTS) for various navigation schemes, layouts and robots-number, and show an improvement of upto 10\% (25 - 1000 sec) for 500 tasks. Our approach is trained with only single-agent A*, but works with other navigation schemes (SIPP and CBS). Hence it can be decoupled from the navigation system.

\end{enumerate}  

\subsection{Related Works}
Multi-robot task allocation (MRTA) has been widely studied in the robotics and related areas such as taxi-matching~\cite{taxi_comb_opti}, food-delivery~\cite{food_delivery}, search and rescue etc. 
MRTA problems are typically solved using (1) market-based approaches and (2) optimization-based approaches for different environments \cite{market_vs_opti}. Market-based approaches are based on auction systems where each robot bids for a task with a base selling price determined by a broker. The negotiation process is based on market theory, where the team seeks to optimize an objective function based on robots' abilities to perform particular tasks \cite{market_coordination}. These solutions \cite{market_effi,market_robust} can be efficient, robust, and scalable but difficult in terms of designing appropriate cost and revenue functions~\cite{market_drawback}. In contrast, optimization-based approaches outperform market-based approaches in handling large-scale MRTA scenarios (such as $50$ tasks and $15$ robots)~\cite{market_vs_opti}. In \cite{opti_tabu}, the idea is to utilize a hybrid optimization approach combining a tabular search with a random search method. The authors in  \cite{opti_tabu_noise} combine a tabular search with a noising method to solve for static tasks, where tasks are not continuously entering but are predefined. Further, \cite{opti_simultaneous} proposes a simultaneous approach to solve the navigation and task allocation problems for a multi-robot system, where simulated annealing and ant colony optimization methods are applied. 
In contrast to the existing above mentioned approaches, our problem is sequential-lifelong single item task-allocation in warehouse settings and we aim to develop an algorithm that can be easily deployed with any navigation techniques. In this work, many auction and optimisation-based approaches boil down to greedy methods or modifications of them, which form our baselines.

\begin{figure*}
\centering
\includegraphics[width=0.75\textwidth]{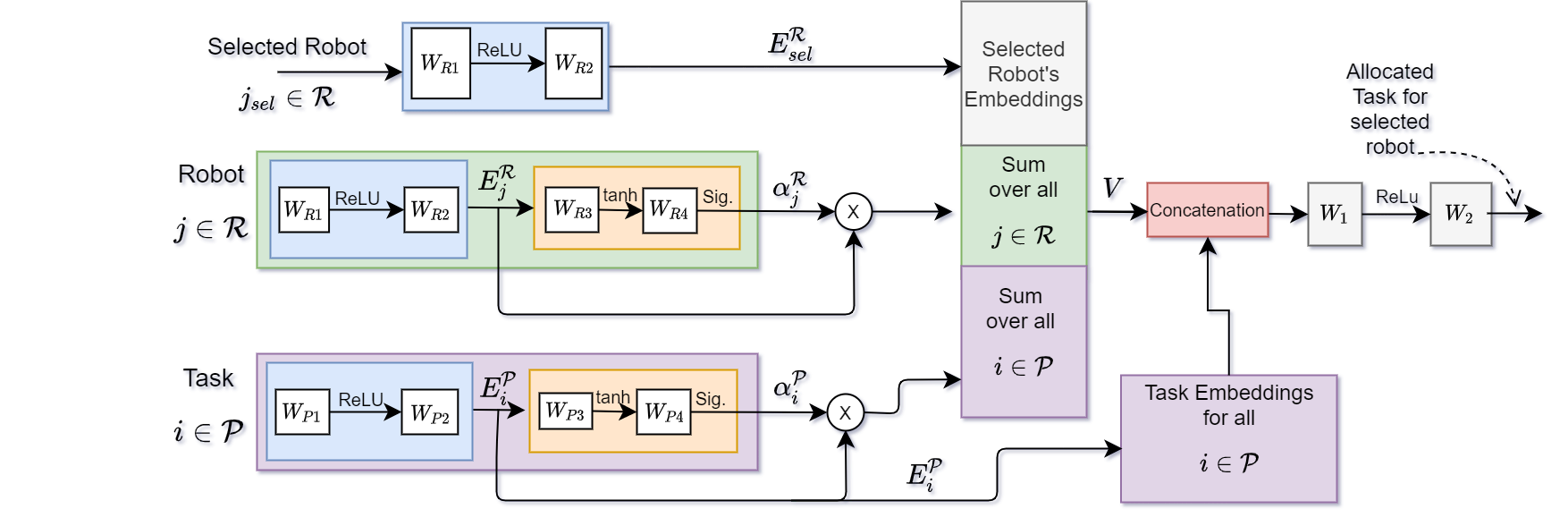}
\caption{We highlight our policy network architecture that is designed to work with variable number of robots and tasks in any warehouse layout. Our formulation takes the feature of robots and tasks as input to generate fixed dimensional embeddings. Next, these embeddings are mixed to generate a global state for the system, which is used to perform the action of choosing tasks for the selected robot. }  
\label{fig:network}
\vspace{-4mm}
\end{figure*}

\textbf{Taxi-Matching Problem:} The problem of task allocation in this work is closely related to the taxi-matching problem, where the goal is to match drivers to customers while reducing the driver wait time, increasing the driver's income, and reducing the driving miles~\cite{inspiration}. The problem of matching pick-up and delivery tasks to drivers was initially solved using combinatorial optimization~\cite{taxi_comb_opti}. Recently, the same problem has been solved via multi-agent RL techniques \cite{ride_share2}, which divide the city map into many dispatching regions (grids) and then learn a policy to assign drivers for a request from the given grid point (state). 

Both taxi-matching and MRTA problems are sequential decision-making problems in terms of allocating tasks. However, warehouses have a fixed and comparatively small layout with a smaller and fixed number of robots deployed. In contrast, ride-sharing platforms tackle thousands of vehicles and millions of orders over a large area like a city. Therefore ride-sharing allocation techniques focus on the distribution of tasks and agents rather than computing their exact coordinate positions.

Thus, the multi-agent formulation for task allocation using reinforcement learning in \cite{ride_share,ride_share2} for taxi-matching applications cannot be directly applied to warehouse settings. 

\textbf{Task Allocation in Warehouses:} Task allocation in warehouses has also been studied~\cite{warehouse_comb}, where it is formulated as an integer linear program and solved using a table-working method. Moreover, \cite{warehouse_hetro_robot} describe an optimal task allocation for heterogeneous robots using navigation path, speed, and current location of each robot and test its validity in simulated environments. These methods are best suited for static settings, where all the robots are available at the same time to solve the matching problem. In practice, different robots take different times to complete a task. Therefore, our goal here is to design dynamic task allocation solutions that assign a task to a robot whenever it becomes available. 

The problem of task allocation in warehouses is also briefly studied in multi-agent pickup and delivery problem. The current state-of-the-art ~\cite{usc} uses greedy solution for task allocation. It continuously allocates and de-allocates tasks based on marginal cost or regret on marginal cost. Such techniques cannot be decoupled from the navigation part. But authors in \cite{mapf_decoupled} proposed a decoupled solution that assigns tasks to agents first in a greedy manner and then utilizes a  Conflict-Based Search (CBS)~\cite{cbs_ta} strategy to plan collision-free paths. We use this commonly used greedy approach (also called MPDM) as one of the baselines.  \\
\noindent

%% file: algo.tex
\vspace{-3mm}
\section{RL Based Formulation}\label{formulation}

{In the existing literature,  multi-robot task allocation problem in warehouse environments is solved by standard optimization methods \cite{task_survey}. But in practice, tasks appear at different time instances, and robots' availability also varies with time, which makes the problem challenging to solve via standard optimization methods. To address this issue, }for the first time in this work, we formulate the multi-robot task allocation problem in warehouse environments as a reinforcement learning problem. The goal is to allocate tasks to multiple robots such that the tasks are performed in an efficient manner (i.e. resources like battery are better utilised and throughput is maximized). The formulation is challenging because it requires to define a Markov Decision Processes (MDP) for the problem. An MDP here is defined by a tuple  $\mathcal{M}:=\left(\mathcal{S}, \mathcal{A}, R, {P}, \gamma\right)$, where $\mathcal{S}$ denotes the finite state space, $\mathcal{A}$ denotes the finite action space, and $R(s,a)$ denotes the reward for state $s$ and action $a$ in $\mathcal{S}$ and $\mathcal{A}$, respectively. Further, ${P}$ represent the transition probability matrix where ${P_a(s,s')}$ is the probability of transition to state $s'\in\mathcal{S}$ from state $s\in\mathcal{S}$ after taking action $a\in\mathcal{A}$.  Here, $\gamma\in(0,1)$ is the discount factor used to trade the near term reward with the long-term rewards in the RL formulation. 

\textbf{MDP model $\mathcal{M}$ for the multi-robot task allocation problem:} Let the time instance be $t\in\mathbb{N}$ and we consider a warehouse environment with the $M$ number of robots collectively denoted by $\mathcal{R}=\{1,2,\cdots,M\}$. A warehouse environment is represented by a specific $K\times K$ grid layout (shown in Fig. \ref{fig:layout_other}). In this work, we do not restrict ourselves to any particular layout, and our approach is applicable to all general layouts. In fact, models trained on dense layouts have shown to give good performance on sparse layouts as well. Our test layout and task generation are described in section 4.
After the task allocation at $t$, the corresponding task is removed from the queue and a new task is added. To make the problem tractable, we assume that at each $t$, only one robot is available to be allocated one of the tasks. We accomplish this by adding a small Gaussian noise of mean $0.25$ and variance $0.25$ to the task completion time. This noise value is chosen to be in (1/100)th range of the task-completion-time range. This ensures that no two robots become available at the same point in time. All the other robots are currently performing the previously allocated tasks. We move from $t$ to $t+1$  as soon as a robot becomes available after completing a task.

\textbf{State:} Each task $i\in \mathcal{P}$ is represented by a tuple $(o_i, d_i, k_i, l_i)$, where $o_i$ is the origin of the task $i$, $d_i$ is the destination of task $i$, $k_i$ denotes the distance between the selected robot's current position to $o_i$, and $l_i$ is the length (distance between $o_i$ and $d_i$) of the task. The origin $o_i$ and destination $d_i$ for each task $i$ are the coordinates sampled from the $2$D layout. Each robot $j\in\mathcal{R}$ is also represented by a tuple $(p_j, r_j)$, where $p_j$ is position ($X-Y$ coordinates) in the $2$D map and $r_j$ is the time left to complete the allocated task (i.e., the task completion time). {Let $j_{sel}$ denotes the available robot for task allocation.} Then, the state $s$ encapsulates the robots' positions, their task completion times, and the task list  $\mathcal{P}$, which consists of tuples $(o_i, d_i, k_i, l_i)$ for all the tasks and the selected robot {$j_{sel}$}. Collectively, we can write the state as $s:=\big\{(p_j, r_j)_{\forall j\in \mathcal{R}}, (o_i, d_i, k_i, l_i)_{\forall i\in \mathcal{P}}, j_{sel} \big\}$ and the collection of all possible states $s$ constitutes the state space $\mathcal{S}$.

\textbf{Action:} An action $a$ in the environment is to select which task to execute from the list of $N$ available tasks in the queue. We define a policy $\pi:\mathcal{S}\rightarrow \mathcal{D}(\mathcal{A})$ (here $\mathcal{D}$ denotes the set of possible distributions defined over $\mathcal{A}$) which takes the current state as input and outputs a distribution across the all possible actions $a\in\mathcal{A}$. Note that in our model, $\mathcal{A}$ constitutes the list of all available tasks from which we need to select a particular task for the selected robot $j_{sel}$. During the training time, we select the action $a\sim \pi(\cdots ~|~ s)$ where  $\pi(\cdot ~|~ s)$ is the probability distribution across the tasks. During the test time, we select the action in a deterministic manner  by choosing a task to accomplish with maximum probability in $\pi(\cdot ~|~ s)$. Further, we discuss the policy architecture in detail in Sec. \ref{architecture}. 

\textbf{Reward:} Designing an appropriate reward for the RL formulation is crucial because this reward eventually decides the behavior of the policy. {We define the reward as the negative of the cost, which is the time a robot takes to reach the origin $o_i$ of the task from its current location. In this work, we use the navigation schemes to calculate the approximate cost of the path by making the robot run through that path.} This makes sense because during the time of travel from the robot's current location to the task origin, the robot is not serving any task, and hence we aim to minimize this time. Specifically, this is defined as total travel delay (TTD), which is the extra time the robot travels when it is not executing any task. Minimizing TTD would ultimately minimize the makespan in the lifelong task allocation. {We note that other reward formulations are also possible (which we tried but didn't work) such as the negative of total \textit{task-length} defined as the time taken to move from the origin to the destination of task. 

Another possibility of reward is the negative of total travel time which is sum of TTD and task-length. That is, the time to travel from robot position to task origin plus time to complete the task. We call it \emph{Reward-Travel Time}. We observe there exists unique minima for TTD. For task-length based rewards, it becomes non-stationary and thus hard to learn. We compare the different reward schemes in Table \ref{tab:reward}. }

\begin{table}
	\centering
		\resizebox{0.8\columnwidth}{!}{
	    \begin{tabular}{|c|c|c|c|c|}
		\hline
		Reward & Equation & Converged & Minimized  & Minimized \\
		& & & TTD & Makespan\\
		\hline
		TTD (proposed) & $\text{dist}(p_j, o_i)$ & Yes & Yes & Yes\\
		Task length & $\text{dist}(o_i, d_i)$ & No & No & No \\
		TTD + Task length & $\text{dist}(p_j, o_i) + \text{dist}(o_i, d_i)$ & Yes & No & No \\
		\hline
	\end{tabular}}
	
	\caption{ The table shows a comparison of different rewards. Our reward (distance between robot position and task origin) minimizes both TTD and makespan and shows good convergence properties. This is not the case for the other two reward formulations, shown in second and third rows.}\label{tab:reward}
	\vspace{-4mm}
\end{table}

\textbf{State-Transitions:} At current state $s_t$, we have a task list in queue $\mathcal{P}$ and one available robot (say $i\in\mathcal{R}$). At this state, we take action $a_t$ according to a policy $\pi(\cdot~|~s_t)$ that decides which task would be allocated to the free robot. The selected task is removed from the list, new task gets added to the list, and the robot becomes unavailable. Our algorithm will wait until next robot becomes available which would mark our new state $s_{t+1}$. Note that the state transitions happens at irregular intervals.

\subsection{Network Architecture using Attention Based Embeddings}\label{architecture}

The objective of an RL problem is to maximize the expected value of the cumulative reward returned under policy $\pi_{\theta}$ which is parametrized by $\theta$. Mathematically, the objective is given as:
\begin{align}\label{main2}
    \max_{\theta} J(\theta),
\end{align}
where $J(\theta) :=V_{\pi_{\theta}}(s_0)$. Here, $V_{\pi_{\theta}}(s_0)$ is the value function for any arbitrary initial state $s_0$ given as 

    $V_{\pi_{\theta}}(s_0)=\mathbb{E}_{\pi}\left[\sum_{t=0}^{\infty}\gamma^t R(s_t,a_t)~\big |~ s_0=s, a_t\sim\pi_{\theta}(\cdot~|~ s_t)\right]$.

The formulation in \eqref{main2} gives rise to class of policy gradient algorithms \cite{zhang2020global}.  The explicit form of the policy gradient is given by 
\begin{align}
    \nabla_{\theta} J(\theta) = ({1}/{1-\gamma}) \mathbb{E}_t[ \nabla_{\theta} \log \pi_{\theta}(a_t~|~ s_t)\cdot A_t],
\end{align}
where $\mathbb{E}_t$ is the empirical average over a finite batch of samples, and $A_t$ denotes the estimate of the advantage function $A_t=Q_{\pi_{\theta_t}}(s_t,a_t) - V_{\theta_t}(s_t)$. We use the popular proximal policy optimization (PPO) algorithm to solve the problem due to its advantages as compared to other RL methods such as data efficiency and reliable performance {\cite{PPO}}. A generalized advantage estimator \cite{schulman2015high} is used to estimate $A_t$ in this work .  Here, $Q_{\pi_{\theta_t}}(s_t,a_t)$ denotes the state action value function. 

\textbf{Attention Inspired Policy Architecture:} To this end, the most important part is to design the policy $\pi_{\theta}$ which is mostly defined as a deep neural network (DNN) in the literature \cite{DBLP:journals/corr/abs-1909-01150} and $\theta$ corresponds to the weights of the DNN. The design of a specific neural network architecture for the policy is crucial and challenging because it will control the class of policies over which we will be optimizing the non-convex objective of \eqref{main2}. A naive architecture would be to simply concatenate all robot and task information (locations, availability etc.) and pass that as input to the network. The input layer size would then depend upon the number of robots and tasks. Moreover, the number of trainable parameters would keep increasing with the increase in number of tasks and robots. This kind of architecture is not practical due to the scalability issues with respect to the number of agents and tasks in the given warehouse environment. 

To address the issue of scalability, we take motivation from the problem of multi-driver vehicle dispatching problem as solved in ~\cite{inspiration} and propose an {attention mechanism inspired policy architecture} shown in Fig. \ref{fig:network}.  The proposed architecture is able to handle arbitrary  number of robots and tasks in any warehouse layout via utilizing global embeddings. For each robot $j \in \mathcal{R}$, let us denote the features $\mathcal{F}_j^{\mathcal{R}}:=\{p_j, r_j\}$, which consists of position coordinates $p_j = (x_{j}, y_{j})$ and time to task completion $r_j$, are passed through one layer of $16$ nodes and ReLU activation. This is followed by an another layer with $16$ nodes to generate robot embeddings denoted by ${E}_j^{\mathcal{R}}$. This is important to extract the right information from the data input in the form of embeddings which is then used for further processing instead of the direct data input. Selected robot's embeddings ${E}^{\mathcal{R}}_{sel}$ are also generated in a similar manner. Mathematically, we have
\begin{equation}
\begin{aligned}\label{first_net}
E_j^{\mathcal{R}} = W_{R2}\cdot ReLu\left(W_{R1}\cdot \mathcal{F}_j^{\mathcal{R}}\right), \\
{E}^{\mathcal{R}}_{sel} = W_{R2}\cdot ReLu\left(W_{R1}\cdot \mathcal{F}_{sel}^{\mathcal{R}}\right).  
\end{aligned}
\end{equation} 
where $W_{R1}$ and $W_{R2}$ are the weights of the neural network. In a similar manner, for each task $i\in\mathcal{P}$, the input features $\mathcal{F}_i^{\mathcal{P}} = (o_i, d_i, k_i, l_i)$ are passed through one layer of $16$ nodes with ReLU activation. This is followed by another $16$ node layer to generate task embeddings ${E}_i^{\mathcal{P}}$ as ${E}_i^{\mathcal{P}} = W_{P2}\cdot ReLu\left(W_{P1}\cdot \mathcal{F}_i^{\mathcal{P}}\right)$.

The robot embeddings ${E}_j^{\mathcal{R}}$ as well as task embeddings ${E}_i^{\mathcal{P}} $ are passed through a $16$ node layer with tanh activation and $1$ node layer with sigmoid activation to compute scalar outputs ${\alpha}_j^\mathcal{R}$ and ${\alpha}_i^\mathcal{P}$, respectively, for a robot and the tasks given by
\begin{align}
    {\alpha}_j^\mathcal{R} = \text{sig}\left(W_{R4}\cdot \text{tanh}\left(W_{R3}\cdot{E}_j^{\mathcal{R}}\right)\right) \\
    {\alpha}_i^\mathcal{P} = \text{sig}\left(W_{P4}\cdot \text{tanh}\left(W_{P3}\cdot{E}_i^{\mathcal{P}}\right)\right)
\end{align}
 These scalar would provide different weight-age to different robots or tasks. 
Next, we compute the weighted average of all robot embeddings as well as task embeddings given by
\vspace{-1mm}
\begin{align}
    v^\mathcal{R} = \sum_j   {\alpha}_j^\mathcal{R}  {E}_j^{\mathcal{R}}, \ \ \text{and} \ \ v^\mathcal{P}  = \sum_i   {\alpha}_i^\mathcal{P}   {E}_i^{\mathcal{P}}.
\end{align} 
\vspace{-6mm}

 Further, the global task embeddings are computed using an attention-like mechanism \cite{attention} so that we generate the expected global information.
The final vectors $v^\mathcal{R}$ and  $v^\mathcal{P}$ are concatenated together with selected robot's embedding ${E}^{\mathcal{R}}_{sel}$ to generate one single vector
$V = (v^\mathcal{R}, v^\mathcal{P}, {E}^{\mathcal{R}}_{sel})$ which denotes the state of the system. We remark that the dimension of this state of the system is independent of the number of tasks or agents in the system which is an important aspect of the proposed architecture. {Moreover, to make an accurate prediction of action, the network requires both global and local information, which motivates us to concatenate selected robot (local embedding) and, $v^\mathcal{R}$ and  $v^\mathcal{P}$ (global embeddings).} Next, this one dimensional vector $V$ is concatenated with two dimensional task embedding using broadcasting, which implies 
 \begin{align}
    \text{Output} = (V, {E}_i^{\mathcal{P}}), \text{ for } \forall i \in \mathcal{P}.
\end{align}
Then the \text{Output} is passed through one layer with $8$ nodes and ReLU activation, then passed through another layer with one node,  and finally passed through categorical distribution to choose the task to be executed.
\begin{align}
   \!\!\!\!\! Task^{chosen} = \text{Categorical}(W_2 \cdot ReLu(W_1 \cdot Output)).
\end{align}
The output ($Task^{chosen}$) is in form of integer denoting the task number the selected robot is assigned to.
We use categorical distribution during training time to sample the chosen task. During test time, we choose the task with the highest probability.

%% file: Simulations.tex
\section{Implementation and Evaluation}\label{experiments}

In this section, we describe our implementation and analyze its performance in different settings by varying the number of robots, task generation schemes, layouts and navigation schemes. During training, we use two different navigation techniques to compute the cost for the RL environment: \emph{direct navigation} and \emph{A*}. In the direct navigation approach, the Euclidean distance between the robot position and task origin is used to evaluate the cost at each instant $t$. In the A* navigation, the A* algorithm is used to calculate the distance. We compare our method (RTAW) with minimum pickup distance minimization (MPDM) and regret based task selection as our baselines. Both these baselines are used in sequential single-item auctions. 

\textbf{MPDM} algorithm chooses the task that is closest to the robot. It is optimal if there is only one robot in the system and the task utility equals the distance of the robot to the task start position (see \cite{taxi_comb_opti} for a more in depth discussion).  MPDM is the only baseline used for comparison in the literature for decoupled task allocation and navigation~\cite{mapf_decoupled}. The other state-of-the-art \cite{usc} solutions for warehouse environments are based on coupled task allocation and navigation. Therefore, they cannot be directly applied. We present a modified version of \textbf{RBTS} used in \cite{usc} inspired from \cite{rbts_1,rbts_2}. 

\textbf{RBTS:}
For every task in the list, find the distance to the closest robot. Subtract the distance of the selected robot to each of these tasks with the previous quantity. Choose the task with the maximum value (i.e. maximum regret). 

For all experiments, we run the test on a fixed task set by fixing the seed of the random number generator and report TTD values for them. We train our model on the  Nvidia GeForce RTX $2080 Ti$ GPU with $11$GB of memory, for $4$ different environments and batch size of $32$. We run the training for $4$ million iterations which takes around $12$ hours with A* navigation algorithm. The network architecture is not bulky with GPU usage less than 5-10 \%. Refer to appendix for the training curves and hyper-parameters tuning. Next, we describe the different experimental settings and discuss the experimental results in detail.  

\subsection{First Experimental Setting}

In appendix sec. \ref{proof-concept}, we show a simple experiment as proof of concept with 2 agents and 5 tasks, to compare the three methods. In appendix sec. \ref{two-task}, we show another simple experiment with two fixed tasks both of which have the same starting point but with different destinations. The two tasks are chosen such that one is always longer than the other. Since the starting coordinates are same, greedy policy will randomly pick a task while RL policy learns to pick the smaller task. Thus, giving $50\%$ and $55\%$ improvement over MPDM for $10$ and $100$ robots, respectively. 

\subsection{Final Experimental Setting}

\begin{figure}[t]
% 	\centering
	\begin{subfigure}{.3\columnwidth}
		\centering
		\includegraphics[trim={0.3cm 0.6cm 0cm 0.5cm, clip},scale=0.26]{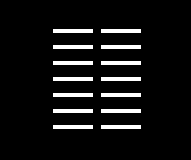}
		\caption{\scriptsize Layout A.}
	\end{subfigure}
	\begin{subfigure}{.3\columnwidth}
		\centering
		\includegraphics[trim={0.3cm 0.6cm 0cm 0.5cm, clip},scale=0.26]{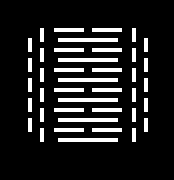}
		\caption{\scriptsize Layout B.}
	\end{subfigure}
	\begin{subfigure}{.3\columnwidth}
		\centering
		\includegraphics[trim={0.3cm 0.6cm 0cm 0.5cm, clip},scale=0.26]{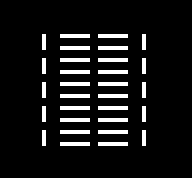}
		\caption{\scriptsize Layout C.}
	\end{subfigure}
	\\
\begin{subfigure}{.4\columnwidth}
		\centering
			\vspace{2mm}
		\ \ \ \includegraphics[trim={0.3cm 0.6cm 0cm 0.5cm, clip},scale=0.26]{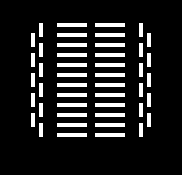}
		\caption{\scriptsize Layout D.}
	\end{subfigure}
	\begin{subfigure}{.4\columnwidth}
		\centering
		\ \ \ \includegraphics[trim={0.3cm 0.6cm 0cm 0.5cm, clip},scale=0.26]{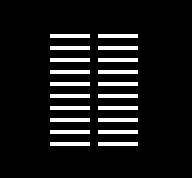}
		\caption{\scriptsize Layout E.}
	\end{subfigure}
	
	\caption{This figure presents different real-world warehouse layouts generated with varying level of compactness using the Asprilo framework for a fixed grid size of $60 \times 60$~\cite{mapd_review, layout2}. We use some of the most challenging layouts. Layout A-very low compactness, Layout B-high compactness, Layout C-low compactness, Layout D-high compactness, Layout E- low compactness.
	}\label{fig:layout_other}
	\vspace{-5mm}
\end{figure}

Next, we raise the difficulty level further and shift focus to a more general setting where the tasks are generated randomly in the environment using designated method.  We report the results in Table \ref{tab:simple_time} for direct navigation and A* navigation, respectively. We use the layouts as shown in \ref{fig:layout_other} for this evaluation.

\textbf{Task Generation:} To the best of our knowledge, there are no publicly available datasets for the warehouse task allocation problem. Therefore, in the literature, the task generation for the experiments is synthetic and has been divided into three different ways random, clustered,  and designated \cite{mapd_review}. For the experiments in this section, we consider designated method for task generation where specific regions for pick-up and delivery are defined and tasks are randomly sampled from them. For our case, we define regions for pick-up and delivery using Gaussian distribution, and randomly sample our tasks from those distributions. This method is the closest to the practical task generation method and is most commonly used in literature \cite{task_gen,task_gen2}.

\begin{table}[h]
	\centering
	\resizebox{1\columnwidth}{!}{%
	\begin{tabular}{|c|c|c|c|c|c|c|} 
		\hline
		\multicolumn{1}{|l|}{}             & Robot No. & MPDM  & RBTS & RTAW & \% Imp. MPDM & \% Imp. RBTS \\%& \% improvement  \\ 
		\hline
		Direct & 10            & 8312 $\pm$ 931  & 8376 $\pm$ 653 & \textbf{7924 $\pm$ 645} & 4.42 $\pm$ 3.27 & 5.36 $\pm$ 3.39  \\%29\%            \\ 
		\cline{2-7}
		& 100           & 13089 $\pm$ 751 & 13322 $\pm$ 905 & \textbf{12298 $\pm$ 747} & 6.04 $\pm$ 1.86 & 7.55 $\pm$ 4.96   \\ %& 7.3\%           \\ 
		\hline
		Layout  & 10 & 5662 $\pm$ 372 & 6194 $\pm$ 235 & \textbf{5165 $\pm$ 132} & 8.77 $\pm$ 4.96 & 16.6 $\pm$ 5.14  \\  
		\cline{2-7}
		 A & 100 & 7429 $\pm$ 213 & 7882 $\pm$ 408 & \textbf{7032 $\pm$ 393} & 5.33 $\pm$ 3.94 & 10.77 $\pm$ 2.84  \\  
		\hline
		Layout  & 10 & 5591 $\pm$ 227 & 6098 $\pm$ 330 & \textbf{4989 $\pm$ 177} & 10.74 $\pm$ 1.28 & 18.1 $\pm$ 3.12 \\  %& 27.8\%          \\ 
		\cline{2-7}
		 B & 100 & 7885 $\pm$ 173 & 8288 $\pm$ 488 & \textbf{7387 $\pm$ 355} & 3.34 $\pm$ 3.22 & 10.7 $\pm$ 5.5 \\  %& 7.3\%             \\
		\hline
		Layout  & 10            & 6205 $\pm$ 247 & 6503 $\pm$ 435 & \textbf{5861 $\pm$ 378} & 5.56 $\pm$ 4.2 & 9.74 $\pm$ 5.24 \\  
		\cline{2-7}
		 C & 100           &  7210 $\pm$ 201 & 7915 $\pm$ 338 & \textbf{6954 $\pm$ 164} & 3.53 $\pm$ 1.69 & 12 $\pm$ 4.63  \\ 
		\hline
		Layout  & 10  & 5920 $\pm$ 501 & 6675 $\pm$ 663 & \textbf{5719 $\pm$ 403} & 3.27 $\pm$ 3.23 & 14.09 $\pm$ 3.96 \\  %& 27.8\%          \\ 
		\cline{2-7}
		 D & 100           & 7764 $\pm$ 764  & 8182 $712$ &  \textbf{7101 $\pm$ 357} & 8.11 $\pm$ 6.22 &  13 $\pm$ 4.56\\  %& 7.3\%             \\
		\hline
		Layout  & 10            & 5536 $\pm$ 367 & 5916 $\pm$ 259 & \textbf{5281 $\pm$ 312} & 4.6 $\pm$ 2 & 10.7 $\pm$ 3 \\ 
		\cline{2-7}
		 E & 100           & 7791 $\pm$ 354 & 8126 $\pm$ 349 & \textbf{7453 $\pm$ 207} & 4.33 $\pm$ 3.45 & 8.28 $\pm$ 5.1 \\ 
		\hline
		
	\end{tabular}
	}
	\caption{ TTD time for $500$ tasks in seconds. We assume the robot moves $1$ unit in $1$ sec. We show results for Direct navigation scheme with no obstacles and for A* navigation scheme on various layouts of grid size $60 \times 60$. We generate results for 5 random seeds and report the mean and std of these 5 runs for each value. For eg. in the first entry of the table, [8312 $\pm$ 931], 8312 is mean of 5 runs and 931 is the std.}
	 	\label{tab:simple_time}
	 	\vspace{-2mm}
\end{table}

\textbf{Layout Generation:} In order to evaluate approach, we  considered real-world warehouse environment layouts to test the performance of our algorithms. These are based on~\cite{mapd_review} and warehouse layouts generated through Asprilo. 
We evaluate the performance of the proposed algorithm on all the layouts in Fig. \ref{fig:layout_other} and compare the performance for $10$ and $100$ robots summarized in Table \ref{tab:simple_time} and \ref{tab:comp_nav}.

The task queue length is $10$ for all cases. As shown in Table \ref{tab:comp_nav}, RTAW outperforms greedy (MPDM) as well as regret-based (RBTS) baselines for all layouts and navigation scenarios.
For CBS \cite{cbs_ta} and SIPP \cite{sipp} navigation schemes, we train the algorithm on A* navigation and test on CBS and SIPP. These navigation algorithm takes a lot of time to solve path therefore training with these navigation schemes is infeasible. For the same limitation of these navigation algorithms, we show results only for 10 agents for various layouts. Our proposed architecture trained on one layout can be efficiently used for a similar layout with different compactness. For instance, trained policy network for layouts A and E can be used interchangeably. This is due to the similarity in task-distribution.
\begin{table}[h]
	\centering
		\resizebox{1\columnwidth}{!}{%
	\begin{tabular}{|c|c|c|c|c|c|c|}
		\hline
		Robot Number & Grid Size & MPDM  & RBTS & RTAW & MPDM (\%) &  RBTS (\%) \\
		\hline
		500 & Direct & 547538 & 554681  & \textbf{503530} & 8.04 & 9.22\\
		\hline
		500 & A* on Layout B & 311403 & 301221 & \textbf{278590} & 10.54 & 7.5 \\
		\hline
		1000 & Direct & 695816 & 681236 & \textbf{674089} & 3.12 & 1.04\\
		\hline
		1000 & A* on Layout B & 413406 & 423368 & \textbf{403112} & 2.49 & 4.78 \\
		\hline
	\end{tabular}
		}	\caption{TTD time in seconds for $5000$ tasks for $500$ and $1000$ robots for direct and A* navigation scheme. We assume the robot moves $1$ unit in $1$ sec. Here we show output for only 1 run because with such large number of robots, baseline algorithm (RBTS) takes a very long to run.}
		\label{tab:scalable}
		\vspace{-8mm}
\end{table}

\begin{table}[H]
	\centering
	\resizebox{1.0\columnwidth}{!}{%
	\begin{tabular}{|c|c|c|c|c|c|} 
	    \hline
	     & \multicolumn{5}{|c|}{SIPP} \\   
		\hline
		Layout & MPDM  & RBTS & RTAW & MPDM (\%) & RBTS (\%) \\ 
		\hline
		 A & 692 $\pm$ 106 & 725 $\pm$ 94 & \textbf{645 $\pm$ 67} & 6.7 $\pm$ 6.4 & 11 $\pm$ 4\\ 
		\hline
		 B & 616 $\pm$ 33.7 & 636 $\pm$ 102 & \textbf{566 $\pm$ 70} & 8.4 $\pm$ 7.6 & 10.5 $\pm$ 7.6 \\
		 C & 684 $\pm$ 110 & 707 $\pm$ 138 & \textbf{626 $\pm$ 86} & 8.1 $\pm$ 4.95 & 10.6 $\pm$ 5.5 \\ 
		\hline
		D  & 710 $\pm$ 131 & 706 $\pm$ 165 & \textbf{642 $\pm$ 140.5} & 10 $\pm$ 4.28 & 8.6 $\pm$ 4.7 \\ 
		\hline
		E  & 643 $\pm$ 62.5 & 689 $\pm$ 110 & \textbf{595 $\pm$ 84} & 7.53 $\pm$ 5.5 & 13.7 $\pm$ 8.9 \\
		\hline
		
		 & \multicolumn{5}{|c|}{CBS} \\   
		\hline
		Layout & MPDM  & RBTS & RTAW & MPDM (\%) & RBTS (\%) \\
		\hline
		 A & 644 $\pm$ 80 & 696 $\pm$ 63 & \textbf{615 $\pm$ 74} & 4.5 $\pm$ 1.6 & 11.6 $\pm$ 7.43\\ 
		\hline
		 B & 680 $\pm$ 114 & 675 $\pm$ 105 & \textbf{619 $\pm$ 99} & 8.9 $\pm$ 1.8 & 8.3 $\pm$ 3.7 \\  
		\hline
		 C & 655 $\pm$ 28 & 670 $\pm$ 45 & \textbf{625 $\pm$ 26.4} & 4.5 $\pm$ 3.8 & 6.6 $\pm$ 3.6 \\ 
		\hline
		D  & 766 $\pm$ 121 & 817 $\pm$ 128 & \textbf{738 $\pm$ 125} & 3.73 $\pm$ 2.78 & 9.74 $\pm$ 4.3 \\ 
		\hline
		E & 618 $\pm$ 124 & 671 $\pm$ 127 & \textbf{591 $\pm$ 109} & 4.3 $\pm$ 4.2 & 11.98 $\pm$ 1.7 \\
		\hline
	\end{tabular}
	}

	\caption{TTD time for $50$ tasks (in seconds) for 10 robots for SIPP and CBS navigation scheme respectively. We assume the robot moves $1$ unit in $1$ sec. The model is trained on only A* navigation scheme for various layouts (refer Fig. \ref{fig:layout_other}) of grid size of $60 \times 60$.} 
	 	\label{tab:comp_nav}

\end{table}

\vspace{-4mm}
\subsection{Scalability}\label{scalability}
We show that RTAW is scalable by performing experiments on $500$ and $1000$ robots. For direct navigation, we used a $300\times300$ grid layout with no obstacles. For A* navigation we used a $256 \times 256$ grid size of layout C. The results are summarized in Table \ref{tab:scalable}. We note that RTAW outperforms the baselines even for higher number of robots and tasks.

%% file: conclusions.tex
\section{Analysis and Limitations}

Our final reward function based on TTD exhibits better  behavior than previous reward functions based on task-length. This is because task-length is a fixed quantity and does not aid in learning. It makes it hard to compute the optimal function, whereas TTD has shown to exhibit good converging behavior.
We observe that we do not get good improvement for all the seeds values. There are some cases,  when our model performs similar to greedy strategy. But such cases are not frequent, as shown in our results. When we compute an average over multiple runs,
we always see a clear improvement. 
Some hyper-parameters played an important role in formulating the problem and the performance of the learning method. We choose a high $\gamma$ value as close as possible to 1 because we want to take actions that result in higher future reward and not emphasise only the immediate reward (which happens in the greedy method). We used entropy regularisation to aid the learning and deal with complex environment (i.e., larger robots, more tasks, compact layout). We reduce the entropy coefficient to help in the learning process.  More details are given in the supplementary material.

Our RTAW algorithm has some limitations. One issue is handling the variable load capacity of the robots. Our current approach makes a assumption that a robot executes only one task (going from point A to point B) at a time. Another assumption we make is that robots always finish their task and are ready in terms of performing the next task right away. But this may not hold in practice due to robot's state, battery charging etc. Our current approach can handle variable number of robots so in-order to accommodate new robots coming in and leaving the system, we just need to maintain separate system that keeps track of the total number of robots. Currently, we assume that the robots are similar or homogeneous, that is, all robots travel with same velocity and acceleration, but this is may not be the case always in the real world. 
%In real world scenarios, robots can be heterogeneous. 

\vspace{-1mm}
\section{Conclusion}
\vspace{-1mm}

We considered a problem of multi-robot task allocation (MRTA) in warehouse environment and proposed a novel and state-of-the-art RL-based solution we called RTAW. RTAW is general and can be applied to arbitrary warehouse layouts. We have evaluated the performance of RTAW on multiple layouts and under different experimental settings, where the number of robots is varied from $10$ to $1000$. We observe that our method outperforms prior techniques based on greedy and regret-based strategies and the computational complexity scales linearly with the number of robots. Furthermore, RTAW can be combined with any centralized or decentralized navigation approach.

%In this work, we also could not provide a theoretical bound on improvement and no theoretical guarantee on how much improvement in time will occur. In-fact in some instance we don't see any improvement and the method performs just as good as greedy (MPDM). But for multiple runs we do get significant improvement on an average. 

%% file: appendix.tex
\onecolumn
\section*{\centering Appendix}\label{Appendix}

\section{Experimental Settings}
\subsection{Proof of concept}
\label{proof-concept}
We devise a simple experiment to validate the proof of concept which means that RL based policies are indeed useful as compared to the greedy and regret-based baseline. We consider $2$ robots and $5$ fixed tasks in a 2D map (as listed in Table \ref{tab:task_list}) to be allocated, with a task queue length of $2$. Since there are only $5$ tasks, we record the five step transitions in the environment and report the task allocation for the greedy and regret-based method in Table \ref{tab:greedy_task}. %

\begin{figure}[t]
\centering
	\begin{subfigure}{.5\columnwidth}
		\centering
		\includegraphics[scale=0.2]{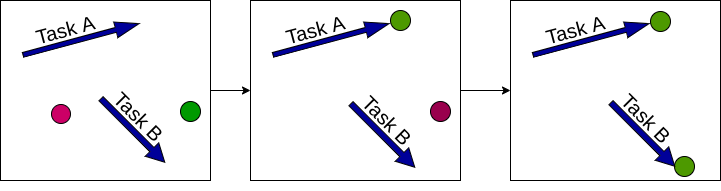}
		\caption{Actions by  RTAW.}
		\label{RL_policy}
	\end{subfigure}\\
\vspace{5mm}
\begin{subfigure}{.5\columnwidth}
		\centering
		\includegraphics[scale=0.2]{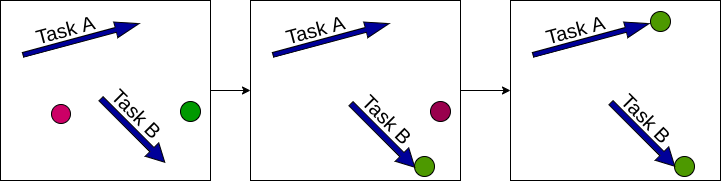}	
			\caption{Actions by greedy baseline.}
			\label{Greedy_policy}
	\end{subfigure}
	\vspace{-2mm}
	\caption{ This figure describes a situation where the proposed RL based policy will take a better decision than the greedy baseline.  Red and green are two robots where the red robot is available for task allocation. Green robot is currently serving a task, and its location is the destination of the task being served. From Fig \ref{RL_policy}, we see that the RL policy chooses Task A over Task B even though Task B is closer. Choosing Task B will give an instant high reward; but since the green robot is too far from Task A it will give a very low reward if it chooses Task A. Thus in order to get higher cumulative reward Task B is selected before Task A. Thus, an RL-based solution results in intelligent task selection which is not possible for a greedy policy as shown in Fig. \ref{Greedy_policy}.}
	\label{fig:rl_intution}
\end{figure}

\begin{table}[H]
	\centering
	\resizebox{0.4\columnwidth}{!}{%
\begin{tabular}{|c|c|l|l|l|l|}
\hline
      & Task 1 & Task 2 & Task 3 & Task 4 & Task 5 \\ \hline
$o_i$ & (2, 9) & (4, 4) & (5, 6) & (2, 4) & (0, 5) \\ \hline
$d_i$ & (5, 5) & (0, 0) & (3, 2) & (1, 2) & (7, 1) \\ \hline
\end{tabular}
% \begin{tabular}{|l|l|l|}
% 			\hline
% 			Task Number  & $o_i$ & $d_i$\\
% 			\hline
% 			Task 1 & (2, 9) & (5, 5) \\
% 			Task 2 & (4, 4) & (0, 0) \\
% 			Task 3 & (5, 6) & (3, 2) \\
% 			Task 4 & (2, 4) & (1, 2) \\
% 			Task 5 & (0, 5) & (7, 1) \\
% 			\hline
% 		\end{tabular}
	}
	\vspace{2mm}
	\caption{We enlist the task list with $5$ tasks with their origin $o_i$ and destination $d_i$ points in the 2D grid map.}
	\label{tab:task_list}
	\vspace{-0mm}
\end{table}
\begin{table}[H]
	\centering
	\resizebox{0.55\columnwidth}{!}{%
		\begin{tabular}{|l|l|l|l|l|l|l|}
			\hline
			State & $(p_i, r_i)$  & Task & TTD & $(p_i, r_i)$  & Task & TTD\\
			\hline
			$S_{0}$ & \textbf{(2, 2, 0)} & Task 1 & 2.8 &  \textbf{(2, 2, 0)} & \textbf{Task 1} & 7     \\
			      &  (6, 2, 2)         & \textbf{Task 2} &  & (6, 2, 2)         & Task 2 & \\
			\hline
			$S_{1}$ & (0, 0, 6.48) & Task 1 & 4.12  &     (5, 5, 10) & \textbf{Task 2} & 2.848\\
			      &  \textbf{(6, 2, 0)}   & \textbf{Task 3} &  & \textbf{(6, 2, 0)}   & Task 3 & \\
			\hline
			$S_{2}$ & \textbf{(0, 0, 0)} & Task 1 &    4.47  & (5, 5, 1.51) & Task 3 & 4.47 \\
			      &  (3, 2, 2.12)         & \textbf{Task 4} & & \textbf{(0, 0, 0)} & \textbf{Task 4} & \\
			\hline
			$S_{3}$ & (1, 2, 4.6) & Task 1 &   4.2  & \textbf{(5, 5, 0)} & \textbf{Task 3} & 1 \\
			      &  \textbf{(3, 2, 0)}         & \textbf{Task 5} &  & (1, 2, 5.2) & Task 5 &\\
		    \hline
			$S_{4}$ & \textbf{(1, 2, 0)} & \textbf{Task 1} &   7.1 & (1, 2, 5.2) & Task 5 & 3.16 \\
			\hline
			\multicolumn{3}{|c|}{Total TTD-Greedy} & 22.69 & \multicolumn{2}{|c|}{Total TTD-Regret TS} & 18.46 \\
			\hline
		\end{tabular}
		}
	\vspace{2mm}
	\caption{ This table shows tasks allocated (in bold) and instant reward (TTD) received at each state for the greedy and regret based baseline.}
	\vspace{0mm}
	\label{tab:greedy_task}
\end{table}
\begin{table}[H]
	\centering
	\resizebox{0.3\columnwidth}{!}{%
		\begin{tabular}{|l|l|l|l|}
			\hline
			State & $(p_i, r_i)$  & Task & TTD\\
			\hline
			$S_{0}$ & \textbf{(2, 2, 0)} & \textbf{Task 1} & 7      \\
			      &  (6, 2, 2)         & Task 2 &  \\
			\hline
			$S_{1}$ & (5, 5, 10) & \textbf{Task 3} & 4.12    \\
			      & \textbf{(6, 2, 0)}   & Task 2 &  \\
			\hline
			$S_{2}$ & (5, 5, 2) & \textbf{Task 4} &  2.24    \\
			      & \textbf{(3, 2, 0)}  & Task 2 &  \\
			\hline
			$S_{3}$ & \textbf{(5, 5, 0)} & Task 5 &   1.41   \\
			      &  (1, 2, 2.3)         & \textbf{Task 2} &  \\
		    \hline
			$S_{4}$ & \textbf{(1, 2, 0)} & \textbf{Task 5} &   3.16 \\
			\hline
			\multicolumn{3}{|c|}{Total TTD} & 18 \\
			\hline
		\end{tabular}
	}
	\vspace{2mm}
	\caption{ This table shows tasks allocated (in bold) and instant reward (TTD) received at each state for our method (RTAW).}
	\label{tab:proposed_task}
\end{table}
Note that the final cost (TTD) incurred by greedy method is $22.69$ and by regret-based method is 18.46. Our task allocation method is presented in Table \ref{tab:proposed_task}, which takes different actions and thereby results in lower TTD of 18. The two robots are initially spawned at locations $(2,2)$ and $(6,2)$  in a 2D grid map of $10\times 10$ and the first robot is available. Our method chooses task $1$ instead of task $2$ in state $s_{0}$ even though choosing task $2$ would have resulted in lower cost. Similarly, our method chooses task $3$ instead of task $2$ in state $s_{1}$ even though task $2$ would have resulted in a better TTD. Overall, we see that the TTD is better for our proposed method because it chooses intelligent action by considering the effect of current action at the future state action transitions. Fig. \ref{fig:rl_intution} pictorially shows an example of the intuition behind the improved performance of our proposed RL-policy over greedy baseline.

\subsection{A Simple Two Task Setting}
\label{two-task}
\begin{table}[H]
	\centering
	\resizebox{0.4\columnwidth}{!}
	{\begin{tabular}{|c|c|c|c|c|} 
	\hline
	\multicolumn{1}{|l|}{}             & Robot No & MPDM   & RTAW & \% Imp.  \\ 
	\hline
	Direct & 10            & 4183 & \textbf{2097}  & 49.86            \\ 
	\cline{2-5}
	& 100           &  4183 & \textbf{1861} & 55.51           \\ 
	\hline
    \end{tabular}}
	\caption{ This table reports the $500$ task completion time in seconds for \emph{Direct} navigation scheme for first experiment with two fixed tasks. The robots moves $1$ unit in $1$ sec in the environment. We note that the the proposed RL based technique works significantly better than the existing MPDM method. We show results for single seed because there is no randomness here.}
	\label{tab:simple_setting}
	\vspace{-8mm}
\end{table}   

In this section, we increase the difficulty level and consider continuously arriving tasks in sets of two, both of which have the same starting point but with different destinations in a $2$D map. The two tasks are chosen such that one is always longer than the other. We specifically choose the tasks: Task A $(0,0,3,3)$ and Task B $(0,0,9,9)$, where $(0,0)$ denotes the origin coordinate while $(3,3)$ and $(9,9)$ are the destination coordinates. All the robots start off at random initial locations in the grid. Since the starting coordinates are same, greedy policy will randomly pick a task. Our intuition behind the RL policy is that it will learn to pick up Task A $(0,0,3,3)$ since the destination coordinate $(3,3)$ will be closer to origin coordinate for future task allocation. The Table \ref{tab:simple_setting} summarises the results and show that RL policy learns to take action corresponding to Task A and minimises the TTD. The results are shown for $10$ and $100$ robots on a plain $10 \times 10$ grid with direct navigation scheme.

\section{Time and Space Complexity of the Proposed Method}\label{time_space}
\begin{figure}[H]
	\centering
	\begin{subfigure}{.5\columnwidth}
		\centering
		\includegraphics[scale=0.28]{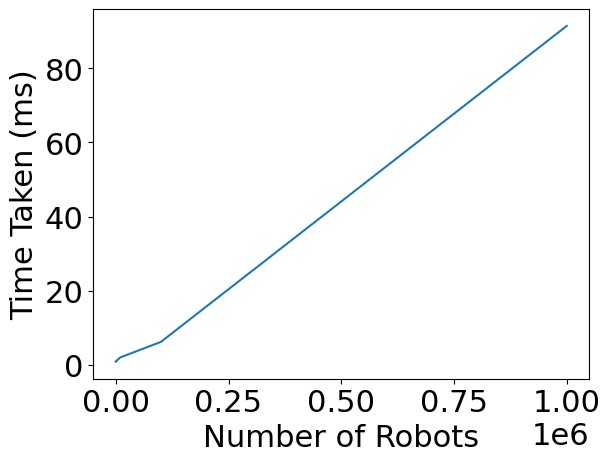}
		\caption{vs No. of Robots.}
	\end{subfigure}%
\hspace{-3cm}	\begin{subfigure}{.5\columnwidth}
		\centering
		\includegraphics[scale=0.28]{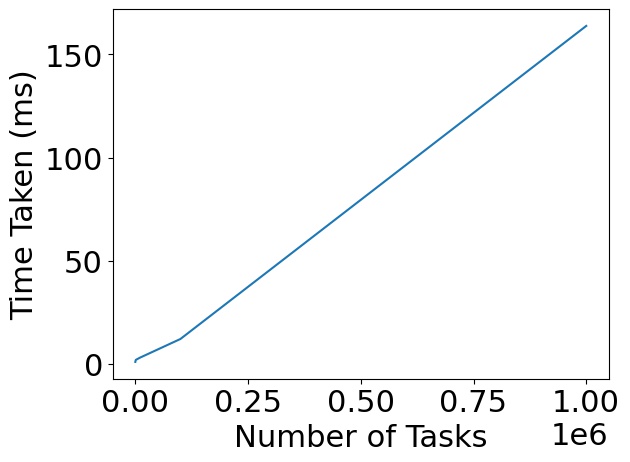}	
			\caption{vs No. of Tasks.}
	\end{subfigure}
	\caption{Scaling of model execution time as a function of number of robots and tasks.}
	\label{fig:time_complex}
\end{figure}

%The time analysis of our proposed method is as shown in fig. \ref{}.
We analyze the performance of our method for one task allocation. The time taken for one forward pass of the model is plotted as a function of number of robots (keeping number of tasks fixed) in Figure. \ref{fig:time_complex}(a). The time taken for one forward pass of the model is plotted as a function of the number of tasks (keeping number of robots fixed) in \ref{fig:time_complex}(b).
From the plots, we see that our method is scalable and linear in both number of agents and number of tasks. Our baseline greedy method is liner but regret-based is exponential.  

The number of weights parameters do not change with increase in the number of the agents or tasks. Hence, it is constant in space complexity.
% Also, solving the problem through combinatorial optimization is NP-hard and the computational cost exponentially increases with number of agents. On the other hand computational cost of one forward pass through our network does not change significantly with the number of agents.

% The ST-SR-IA problem can be formulated as integer linear programming problem with $m$ robots, $n$ prioritized tasks, and utility estimates for each of the $mn$ possible robot-task pairs, we assign at most one task to each robot. A centralized linear programming approach (e.g., Kuhn’s (1955) Hungarian method) will find the optimal allocation in $O(m n^{2})$ time. Since $m = 1$ as we define the problem as sequential task allocation, time complexity changes to $O(n^{2})$.
\section{Training Graphs}
%\vspace{-10mm}
We have included the training curves for the cumulative reward return for the proposed method (RTAW) under different experimental settings considered in the main body of this work. The reward is received after sampling the action from a categorical distribution during training. After every $2k$ time steps we evaluate the model $20$ times and plot the mean and median graph of reward. In the following figures, y-axis denotes the cumulative reward return and x-axis denotes the number of episodes. Hyper-parameters used for training are mentioned in the Table \ref{tab:hyperparameter}.

\begin{table}[H]
	\centering
	\resizebox{0.35\columnwidth}{!}{%
		\begin{tabular}{|l|l|}
			\hline
		    Optimizer  & Adam \\
			Learning Rate & 3e-4 \\
			Update Interval while training & 512 \\
			Epoch per iteration of PPO & 16 \\
			Gamma ($\gamma$) & 0.99 \\
			Lambda ($\lambda$) & 0.95 \\
			Entropy coefficient & 0.01 - 0.001\\
			Value function coefficient & 0.0002 \\
			Policy function coefficient & 0.02 \\
			\hline
		\end{tabular}
	}
	\vspace{2mm}
	\caption{This table summarizes the hyper-parameters used for the training of proposed network architecture via proximal policy optimization (PPO) algorithm \cite{PPO}.}.
	\label{tab:hyperparameter}
\end{table}

\begin{figure}[H]
	\centering
	\begin{subfigure}{.3\columnwidth}
		\centering
		\includegraphics[clip,scale=0.5]{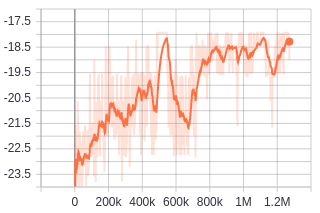}
		\caption{}
	\end{subfigure}%
	\hspace{3mm}
	\begin{subfigure}{.3\columnwidth}
		\centering
		\includegraphics[clip,scale=0.5]{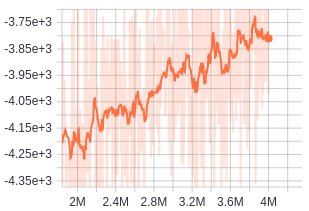}
		\caption{}
	\end{subfigure}%
		\hspace{2mm}
	\begin{subfigure}{.3\columnwidth}
		\centering
		\includegraphics[clip,scale=0.5]{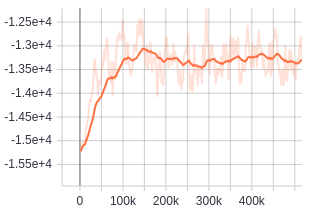}
		\caption{}
	\end{subfigure}\\
	\begin{subfigure}{.35\columnwidth}
		\centering
		\includegraphics[clip,scale=0.5]{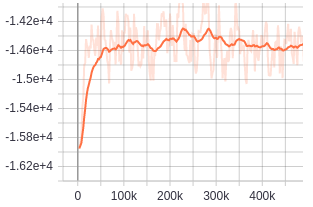}
		\caption{}
	\end{subfigure}%
	\begin{subfigure}{.35\columnwidth}
		\centering
		\includegraphics[clip,scale=0.5]{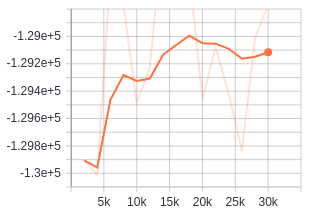}
		\caption{}
	\end{subfigure}
	\caption{(a) Proof of concept reward for 2 agents and 5 tasks. (b) Reward for the second experimental setting with 100 agents. (c) Reward for final experimental setting for 10 agent on 64x64 grid of Layout A. (d) Reward for 100 agent training on 64x64 grid of layout A. (e) Reward for 1000 agent training on 64x64 grid of Layout A.}
	\label{fig:reward_all}
\end{figure}

\begin{figure}
	\centering
	\begin{subfigure}{.3\columnwidth}
		\centering
		\includegraphics[clip,scale=0.5]{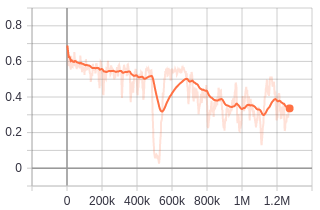}
		\caption{}
	\end{subfigure}%
	\hspace{3mm}
	\begin{subfigure}{.3\columnwidth}
		\centering
		\includegraphics[clip,scale=0.5]{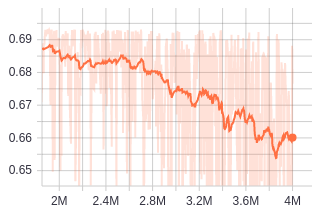}
		\caption{}
	\end{subfigure} %
	\begin{subfigure}{.3\columnwidth}
		\centering
		\includegraphics[clip,scale=0.5]{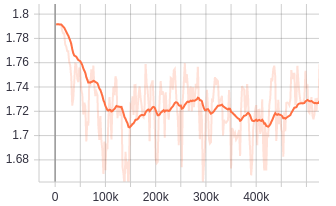}
		\caption{}
	\end{subfigure}\\
	\begin{subfigure}{.35\columnwidth}
		\centering
		\includegraphics[clip,scale=0.5]{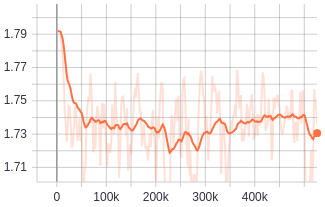}
		\caption{}
	\end{subfigure}%
	\begin{subfigure}{.35\columnwidth}
		\centering
		\includegraphics[clip,scale=0.5]{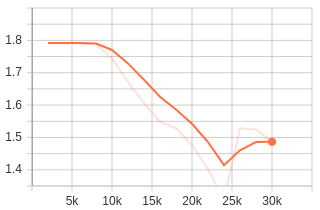}
		\caption{}
	\end{subfigure}
	\caption{Entropy plots. (a) Proof of concept reward for 2 agents and 5 tasks. (b) Reward for the second experimental setting with 100 agents. (c) Reward for final experimental setting for 10 agent on 64x64 grid of layout A. (d) Reward for 100 agent training on 64x64 grid of layout A. (e) Reward for 1000 agent training on 64x64 grid of layout A.}
	\label{fig:reward_all}
\end{figure}